\documentclass[runningheads]{llncs}
\usepackage[T1]{fontenc}
\usepackage{graphicx}
\usepackage[year=2025,ID=122]{iciap}
\usepackage{graphicx}
\usepackage{caption}
\usepackage{subcaption}
\usepackage{tikz}
\usepackage{pgfplots}
\usetikzlibrary{spy,calc}
\usepackage{hyperref}
\usepackage{amsmath}
\usepackage{dblfloatfix}
\usepackage{float} 
\usepackage{booktabs}
\usepackage{multirow}

\newcommand{\hide}[1]{}

\begin{document}
\title{Benchmark on Monocular Metric Depth Estimation in Wildlife Setting}
%
%
\author{Niccolò Niccoli\inst{1}\orcidID{0009-0002-1656-3628} \and
Lorenzo Seidenari\inst{1}\orcidID{0000-0003-4816-0268} \and
Ilaria Greco\inst{1}\orcidID{0000-0002-6208-2465} \and
Francesco Rovero\inst{1}\orcidID{0000-0001-6688-1494}
}
\authorrunning{Niccolò Niccoli et al.}
%
\institute{University of Firenze, Italy \\
\email{\{name.surname\}@unifi.it}
}
\maketitle              
\begin{abstract}
Camera traps are widely used for wildlife monitoring, but extracting accurate distance measurements from monocular images remains challenging due to the lack of depth information. While monocular depth estimation (MDE) methods have advanced significantly, their performance in natural wildlife environments has not been systematically evaluated. This work introduces the first benchmark for monocular metric depth estimation in wildlife monitoring conditions. We evaluate four state-of-the-art MDE methods (Depth Anything V2, ML Depth Pro, ZoeDepth, and Metric3D) alongside a geometric baseline on 93 camera trap images with ground truth distances obtained using calibrated ChARUCO patterns. Our results demonstrate that Depth Anything V2 achieves the best overall performance with a mean absolute error of 0.454m and correlation of 0.962, while methods like ZoeDepth show significant degradation in outdoor natural environments (MAE: 3.087m). We find that median-based depth extraction consistently outperforms mean-based approaches across all deep learning methods. Additionally, we analyze computational efficiency, with ZoeDepth being fastest (0.17s per image) but least accurate, while Depth Anything V2 provides an optimal balance of accuracy and speed (0.22s per image). This benchmark establishes performance baselines for wildlife applications and provides practical guidance for implementing depth estimation in conservation monitoring systems.
\end{abstract}

\section{Introduction}\label{sec:introduction}
Monitoring wildlife populations is critical for biodiversity conservation, ecological research, and environmental management. Camera traps have emerged as a widely used tool for non-invasive wildlife observation, capturing large amounts of visual data across diverse habitats and temporal conditions. Modern camera traps utilize infrared sensors to enable continuous monitoring in low-light and nighttime conditions, providing critical insights into nocturnal species behavior without human disturbance. However, extracting accurate ecological information such as animal density or behavior patterns from these images remains a challenging task, especially in the absence of depth information and under the variable illumination conditions that characterize 24-hour wildlife monitoring. Monocular depth estimation (MDE), which infers scene depth from a single image, has the potential to significantly enhance the utility of camera trap imagery by enabling precise distance measurements without requiring stereo setups or additional sensors.

Despite recent advances in deep learning-based MDE, the application of these methods in wildlife monitoring scenarios has been largely unexplored. Natural environments present unique challenges not addressed by existing benchmarks: variable lighting, occlusion from vegetation, motion blur due to animal movement, and scale ambiguity from inter-species size variation. Moreover, the lack of dedicated datasets with accurate depth annotations in outdoor wildlife contexts has limited systematic evaluation and comparison of existing MDE models in these settings.

In this work, we introduce the first benchmark dedicated to monocular metric depth estimation in wildlife monitoring conditions. Using a custom dataset collected with camera traps and annotated using calibrated ChARUCO patterns, we assess the performance of state-of-the-art MDE models in realistic outdoor scenes. Our contributions are as follows:
\begin{itemize}
    \item we provide a novel evaluation protocol tailored to wildlife applications
    \item we compare the performance of leading MDE models on real camera trap imagery with metric ground truth
    \item we analyze the impact of different post-processing strategies for extracting final distance estimates from predicted depth maps
\end{itemize}

\section{Related Works}\label{sec:relatedworks}

\subsection{Monocular Depth Estimation}

Monocular depth estimation has evolved significantly with the advancement of deep learning approaches. Traditional methods relied primarily on geometric principles and known object sizes for distance calculation through perspective projection\cite{aruco}. While effective in controlled environments, these approaches suffer from reduced accuracy in complex natural scenes where object sizes may be unknown or occlusions frequently occur.

Modern deep learning-based approaches have revolutionized the field by providing more automated and scalable solutions. Ranftl et al.\cite{midas} demonstrated that robust monocular depth estimation models require training on multiple datasets, employing multi-objective optimization strategies to enhance generalization across diverse scenarios. This work laid the foundation for current state-of-the-art methods that can handle varied environmental conditions.

Recent advances have focused on achieving both high accuracy and computational efficiency. Bochkovskii et al.\cite{ml-depth-pro} developed ML Depth Pro, which generates highly detailed and metrically accurate depth maps without relying on camera metadata, achieving results in fractions of a second. Similarly, Yang et al.\cite{depth_anything_v2} introduced Depth Anything V2, a more robust monocular depth estimation model built using a data engine that collects and automatically annotates large-scale unlabeled data, subsequently fine-tuned with metric depth information.

Other notable contributions include MiDaS\cite{midas}, which focuses on relative depth estimation with strong generalization capabilities, and ZoeDepth\cite{zoedepth}, which combines relative and metric depth estimation through a unified framework. These methods represent the current state-of-the-art in general-purpose monocular depth estimation.

\subsection{Depth Estimation in Wildlife and Outdoor Settings}

The application of depth estimation to wildlife monitoring presents unique challenges compared to traditional computer vision benchmarks. Most existing benchmarks focus on urban environments (KITTI\cite{kitti}) or indoor scenes (NYU Depth V2\cite{nyu}), with limited representation of natural outdoor environments characterized by dense vegetation, variable lighting, and organic scene structures.

AUDIT\cite{Johanns_2022} represents a particularly relevant contribution, introducing a fully automated pipeline for estimating camera-to-animal distances by aligning relative monocular depth with metric scale, effectively removing the need for reference images. Their approach demonstrates significant advancement in automated distance estimation, though evaluation was conducted primarily in controlled zoo environments rather than natural wildlife habitats.

Automated camera calibration methods have been developed for various applications, but these methods are usually trained on urban and indoor datasets and lack exposure to wildlife imagery, limiting their effectiveness in natural monitoring scenarios. Even methods like DepthAnyCamera\cite{Guo2025DepthAnyCamera}, which can handle diverse scenes, typically require the system to have observed specific objects during training, presenting challenges when applied to wildlife contexts with unpredictable subjects and environments.

The domain gap between training datasets and wildlife applications represents a fundamental challenge. Natural scenes contain complex depth discontinuities from vegetation, highly variable lighting conditions, and scale ambiguity due to the diverse sizes of wildlife subjects. Unlike urban environments with standardized objects that provide implicit scale references, wildlife scenes lack consistent size cues, making depth estimation particularly challenging.

\subsection{Integration of Depth Estimation in Wildlife Applications}

The integration of accurate depth estimation into wildlife monitoring systems enables several important applications beyond simple distance measurement. Precise distance information improves animal density estimation through more accurate calculation of detection areas and reduced double-counting errors. For behavioral analysis, depth information enhances tracking algorithms by providing three-dimensional trajectory information, enabling better understanding of movement patterns and habitat utilization.

Camera trap positioning and coverage optimization also benefits from reliable depth estimation. Understanding the effective detection range and spatial coverage of installations allows researchers to design more efficient monitoring networks and reduce gaps in spatial coverage. Additionally, depth information can improve automated species identification by providing size constraints that help distinguish between species of similar appearance but different body sizes.

However, the computational requirements and domain-specific challenges highlighted in wildlife applications present ongoing research opportunities. The development of wildlife-specific depth estimation methods, domain adaptation techniques, and hybrid approaches combining geometric and learning-based methods represent promising directions for advancing the field.

\section{Evaluated Methods}\label{sec:method-evaluated}

We evaluate several state-of-the-art monocular depth estimation approaches to determine their effectiveness in wildlife monitoring scenarios. Our comparison includes four deep learning-based methods and one geometric approach as a baseline reference.

\subsection{Deep Learning-Based Methods}

\paragraph{Depth Anything V2} We employ Depth Anything V2\cite{depth_anything_v2} with the ViT-L backbone, which represents one of the most recent advances in foundation models for depth estimation. The model is used with weights fine-tuned on the Hypersim dataset\cite{hypersim} and configured with the default scale parameter of $\alpha = 20$ as recommended by the authors. This model is designed to handle diverse scenes through extensive pre-training on multiple datasets.

\paragraph{ML Depth Pro} ML Depth Pro\cite{ml-depth-pro} is evaluated using its default configuration without additional hyperparameter tuning. This method focuses on producing metric depth estimates directly without requiring post-processing scale adjustment, making it particularly suitable for applications requiring absolute depth measurements.

\paragraph{ZoeDepth} For ZoeDepth\cite{zoedepth}, we utilize the implementation available through Hugging Face's Transformers library with default parameters. ZoeDepth is notable for its ability to combine relative and metric depth estimation through a unified framework.

\paragraph{Metric3D} We evaluate Metric3D v2\cite{metric3d} using its \textit{DINO2reg-ViT-giant2} encoder configuration with publicly available pre-trained weights. Metric3D v2 is a geometric foundation model specifically designed for zero-shot metric depth and surface normal estimation from single images. The model employs a canonical space transformation approach that addresses scale and shift variations across different camera setups, making it particularly robust for metric depth prediction without requiring camera intrinsics or post-processing calibration.

All deep learning models are used with their publicly available general-purpose weights, as no wildlife-specific fine-tuned versions were available at the time of evaluation.

\subsection{Geometric Baseline Method}

As a reference approach, we implement a perspective projection-based method using calibrated markers. ChARUCO patterns of known dimensions are placed at measured distances from the camera trap. The geometric approach estimates depth through homography computation between the ground plane and the image plane, assuming local planarity of the scene. Multiple ChARUCO patterns are used to ensure robust localization and homography estimation is performed using RANSAC to handle outliers and improve robustness against noise. For this method, the depth estimation corresponds to the distance of the lower middle point of the bounding box, which represents the contact point between the pattern and the ground plane.

\subsection{Depth Extraction Strategies}

For the deep learning-based methods, we evaluate two approaches for extracting the final depth estimate from the predicted depth maps:

\begin{itemize}
\item \textbf{Median-based extraction}: The depth value is computed as the median of all pixel depths within the target bounding box. This approach is robust to outliers and provides a stable central tendency measure.
\item \textbf{Mean-based extraction}: The depth value is computed as the arithmetic mean of all pixel depths within the bounding box. This approach captures the overall depth distribution but may be sensitive to extreme values.
\end{itemize}

The bounding boxes are manually annotated around the ChARUCO patterns to ensure consistent evaluation across all methods. This dual extraction strategy allows us to assess the impact of aggregation methods on the final depth estimation accuracy.

\section{Experiments}\label{sec:experiments}

\subsection{Dataset}

We evaluate the depth estimation methods on a custom dataset specifically designed to simulate wildlife monitoring conditions. The dataset consists of 93 images captured using camera traps in outdoor natural environments, with human subjects holding ChARUCO patterns at various distances from the camera. This setup allows us to obtain ground truth distance measurements while maintaining the realistic imaging conditions encountered in wildlife monitoring applications.

Each image is manually annotated with two key pieces of information: (1) the precise distance between the camera and the pattern and (2) tight bounding boxes around the patterns for consistent evaluation. The use of camera traps ensures that all optical characteristics, including lens distortion, depth of field, and image quality limitations typical of wildlife monitoring equipment, are accurately represented in our evaluation.

The dataset encompasses multiple outdoor locations to ensure diversity in environmental conditions, including variations in lighting, vegetation density, terrain topology, and background complexity. Distance measurements range primarily from 1 to 5 meters at 1-meter intervals, with some locations extending up to 7 meters to cover the typical range of wildlife detection scenarios.

\begin{figure}[ht]
    \centering
    \begin{subfigure}[b]{0.48\textwidth}
        \includegraphics[width=\textwidth]{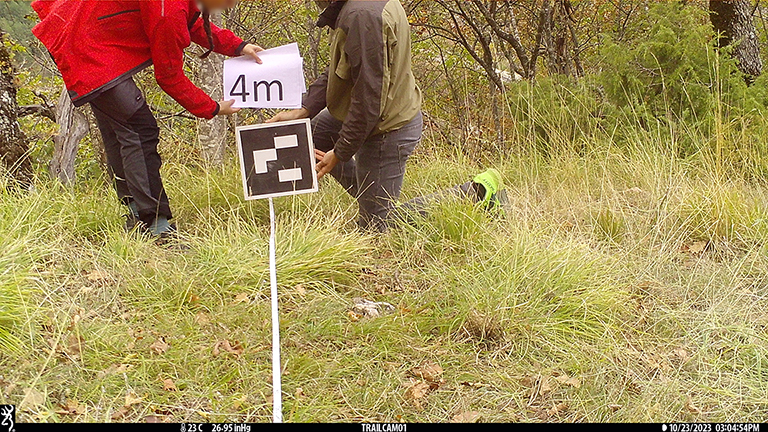}
        \label{fig:sub1}
    \end{subfigure}
    \hfill
    \begin{subfigure}[b]{0.48\textwidth}
        \includegraphics[width=\textwidth]{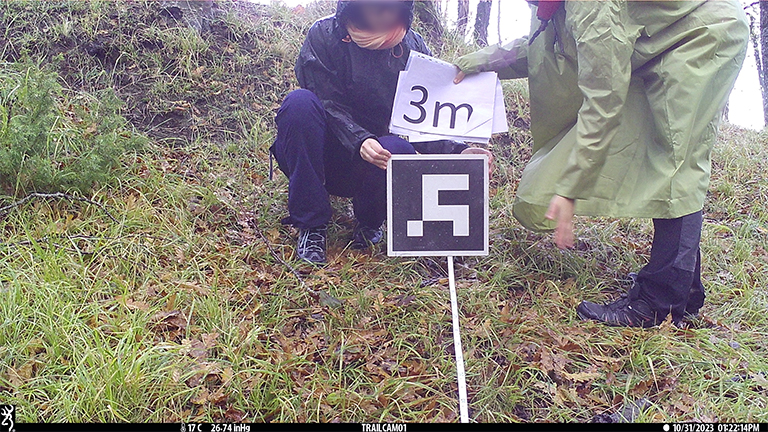}
        \label{fig:sub2}
    \end{subfigure}
    \caption{Representative examples from our evaluation dataset showing the diversity of natural outdoor environments and imaging conditions encountered in wildlife monitoring scenarios.}
    \label{fig:dataset_examples}
\end{figure}

\subsection{Evaluation Metrics}

We assess the performance of each method using four complementary metrics that capture different aspects of depth estimation accuracy:

\begin{itemize}
\item \textbf{Mean Absolute Error (MAE)}: Measures the average absolute difference between predicted and ground truth distances, providing a direct assessment of estimation accuracy in meters.
\item \textbf{Pearson Correlation Coefficient}: Evaluates the linear relationship between predicted and actual distances, indicating how well the method preserves relative depth ordering.
\item \textbf{Relative Error}: Computed as the absolute error normalized by the ground truth distance, offering scale-invariant performance assessment across different distance ranges.
\item \textbf{Root Mean Square Error (RMSE)}: Provides a measure that penalizes larger errors more heavily, revealing the presence of significant outliers in the predictions.
\end{itemize}

We employ both absolute error metrics and correlation measures to provide complementary insights into depth estimation performance. Error metrics (MAE, RMSE, and relative error) quantify the absolute accuracy of distance predictions, measuring how closely estimated values match ground truth measurements in metric units. Correlation analysis evaluates the preservation of spatial depth relationships and scene structure, independent of absolute scale. This dual evaluation approach is particularly valuable because high correlation with moderate absolute errors indicates that a method captures the correct depth structure, suggesting potential for post-processing scale correction to improve metric accuracy.

\subsection{Quantitative Results}

Table \ref{tab:main_results} presents the main quantitative comparison using median-based depth extraction, which consistently showed superior performance across all methods. The results demonstrate clear performance differences between the approaches, with Depth Anything V2 achieving the best overall performance across most metrics.

\begin{table}[h!]
    \centering
    \begin{tabular}{lcccc}
        \toprule
        Method & MAE ($\downarrow$) & Correlation ($\uparrow$) & Rel. Error ($\downarrow$) & RMSE ($\downarrow$) \\
        \midrule
        Depth Anything V2\cite{depth_anything_v2} & \textbf{0.454} & 0.962 & \textbf{0.211} & \textbf{0.593} \\
        ML Depth Pro\cite{ml-depth-pro} & 1.127 & 0.931 & 0.336 & 1.387 \\
        ZoeDepth\cite{zoedepth} & 3.087 & 0.625 & 1.068 & 4.038 \\
        Metric3D\cite{metric3d} & 0.867 & \textbf{0.974} & 0.285 & 0.998\\
        Projection-based & 0.505 & 0.900 & 0.282 & 0.697 \\
        \bottomrule
    \end{tabular}
    \caption{Main quantitative comparison using median-based depth extraction. Bold values indicate the best performance for each metric. All distance measurements are in meters.}
    \label{tab:main_results}
\end{table}

Depth Anything V2 achieves the lowest MAE of 0.454m and the lowest relative error of 0.211, demonstrating superior accuracy in distance estimation. However, Metric3D achieves the highest correlation of 0.974, indicating excellent preservation of depth structure and relationships. Both methods significantly outperform ML Depth Pro and ZoeDepth in this outdoor wildlife setting. The geometric ChARUCO pattern method provides competitive performance (MAE: 0.505m, Correlation: 0.900), serving as a strong baseline that validates the quality of our evaluation setup. Notably, ZoeDepth shows the poorest performance across all metrics, with significantly higher errors that may indicate limited optimization for outdoor natural environments.

To analyze the impact of different depth extraction strategies, Table\ref{tab:extraction_comparison} compares median and mean-based approaches for the deep learning methods only.

\begin{table}[h!]
    \centering
    \resizebox{\textwidth}{!}{%
    \begin{tabular}{lcccccccc}
        \toprule
        \multirow{2}{*}{Method} & \multicolumn{4}{c}{Median Extraction} & \multicolumn{4}{c}{Mean Extraction} \\
        \cmidrule(lr){2-5} \cmidrule(lr){6-9}
        & MAE ($\downarrow$)& Corr. ($\uparrow$)& Rel. Err. ($\downarrow$)& RMSE ($\downarrow$)& MAE ($\downarrow$)& Corr. ($\uparrow$)& Rel. Err. ($\downarrow$)& RMSE ($\downarrow$)\\
        \midrule
        Depth Anything V2\cite{depth_anything_v2} & \textbf{0.454} & 0.962 & \textbf{0.211} & \textbf{0.593} & 0.614 & 0.934 & 0.348 & 0.743 \\
        ML Depth Pro\cite{ml-depth-pro} & 1.127 & 0.931 & 0.336 & 1.387 & 0.901 & 0.749 & 0.299 & 1.155 \\
        ZoeDepth\cite{zoedepth} & 3.087 & 0.625 & 1.068 & 4.038 & 3.590 & 0.579 & 1.301 & 4.739 \\
        Metric3D\cite{metric3d} & 0.867 & \textbf{0.974} & 0.285 & 0.998 & 0.616 & 0.933 & 0.210 & 0.775\\
        \bottomrule
    \end{tabular}    }
    \caption{Comparison of median vs. mean-based depth extraction strategies for deep learning methods. All distance measurements are in meters.}
    \label{tab:extraction_comparison}
\end{table}

The comparison reveals that median-based extraction generally outperforms mean-based extraction for most methods and metrics. However, Metric3D shows an interesting exception where mean extraction yields better MAE (0.616 vs 0.867) and relative error (0.210 vs 0.285), suggesting that this method may produce more stable depth predictions with fewer outliers. For the remaining methods, median extraction provides more robust performance, particularly beneficial when dealing with potentially noisy depth predictions in natural outdoor scenes where vegetation, shadows, or other environmental factors may introduce outliers.

\subsection{Computational Efficiency Analysis}

We evaluate the computational requirements of each deep learning method by measuring inference time on identical hardware configurations. Table \ref{tab:speed} presents the runtime performance for processing single images on a NVIDIA RTX 4090.

\begin{table}[h!]
    \centering
    \begin{tabular}{cc}
    \toprule
        Method & Inference Time (seconds) \\
        \midrule
         ZoeDepth\cite{zoedepth} & \textbf{0.17}\\
         Depth Anything V2\cite{depth_anything_v2} & 0.22\\
         ML Depth Pro\cite{ml-depth-pro} & 0.65 \\
         Metric3D\cite{metric3d} & 0.56\\
         \bottomrule
    \end{tabular}
    \caption{Inference time comparison for deep learning-based depth estimation methods. All measurements performed on identical hardware with single image processing.}
    \label{tab:speed}
\end{table}

ZoeDepth demonstrates the fastest inference time at 0.17 seconds per image, followed closely by Depth Anything V2 at 0.22 seconds. Metric3D offers a reasonable balance between speed (0.56 seconds) and accuracy, positioning itself as a viable option for applications requiring both good performance and reasonable computational efficiency. ML Depth Pro requires the most computation time at 0.65 seconds per image. For wildlife monitoring applications where real-time or near-real-time processing may be beneficial, Depth Anything V2 presents the optimal trade-off between accuracy and speed, while Metric3D provides an alternative for scenarios where highest correlation is prioritized over processing speed.

\subsection{Qualitative Analysis}

\begin{figure}[ht]
    \centering
    
    \begin{subfigure}[b]{0.45\textwidth}
        \centering
        \includegraphics[width=\textwidth]{img/cam15rc22_no_bar/ULTRA_RESIZED_censored_cam15rc224m.jpg}
        \caption{Ground Truth}
        \label{fig:img1}
    \end{subfigure}
    
    \vspace{0.8em}
    
    \begin{subfigure}[b]{0.45\textwidth}
        \centering
        \includegraphics[width=\textwidth]{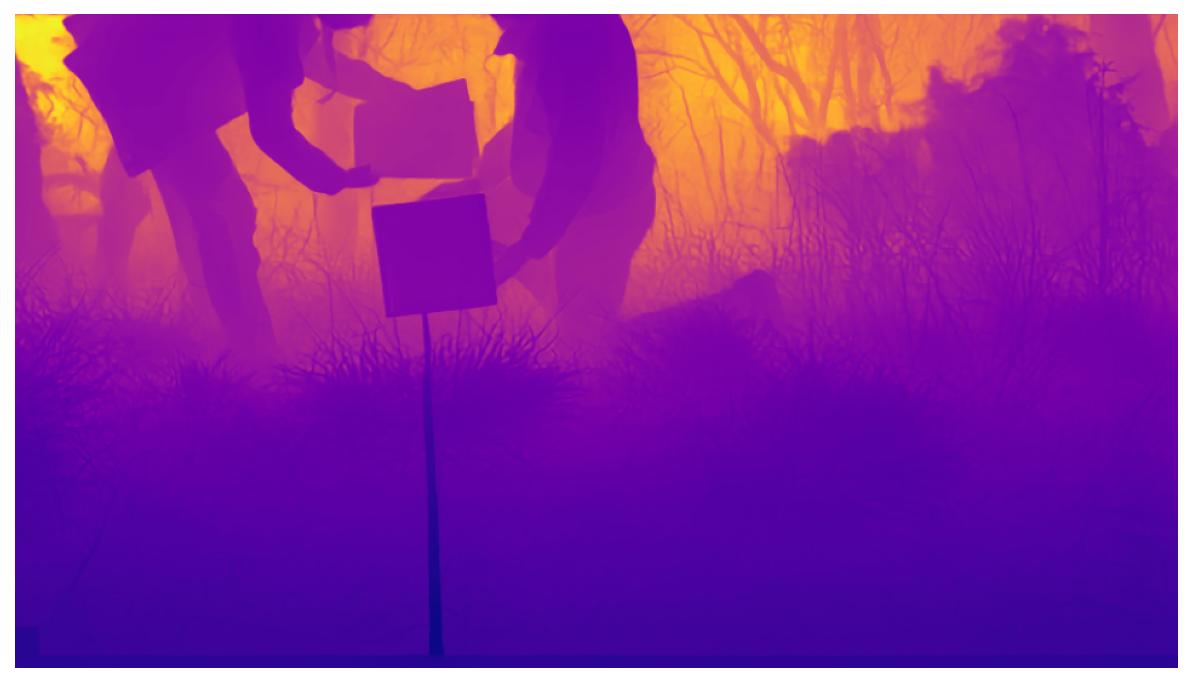}
        \caption{Depth Anything V2}
        \label{fig:img2}
    \end{subfigure}
    \hfill
    \begin{subfigure}[b]{0.45\textwidth}
        \centering
        \includegraphics[width=\textwidth]{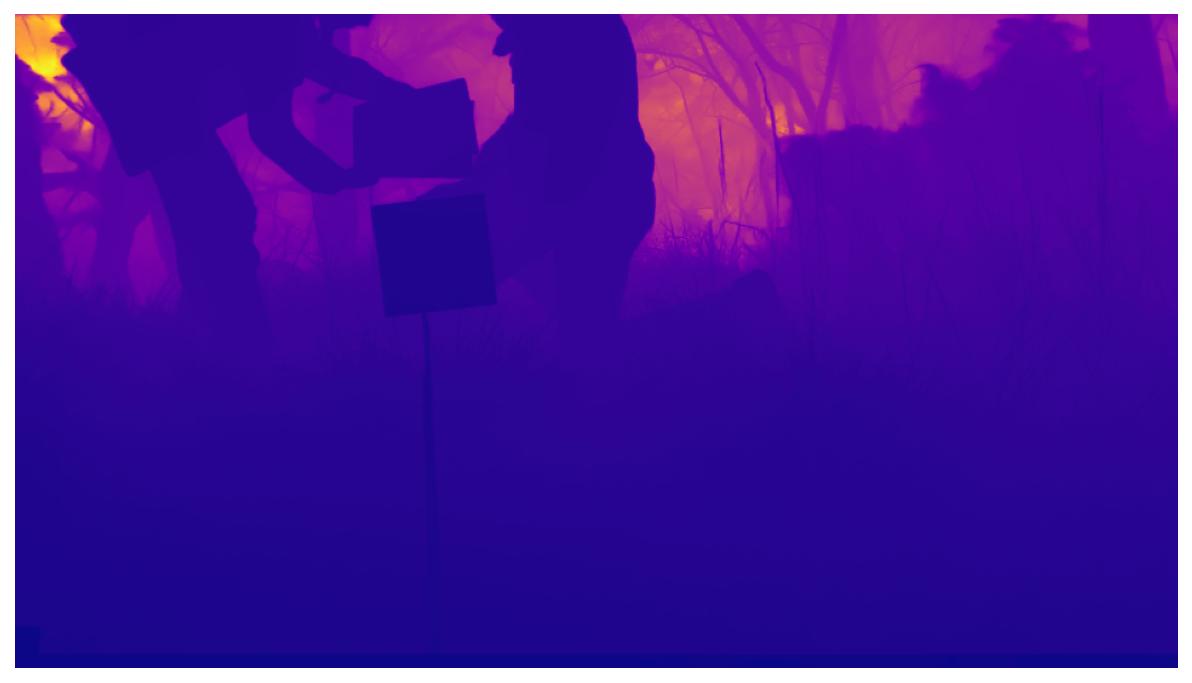}
        \caption{ML Depth Pro}
        \label{fig:img3}
    \end{subfigure}
    
    \vspace{0.5em}
    
    \begin{subfigure}[b]{0.45\textwidth}
        \centering
        \includegraphics[width=\textwidth]{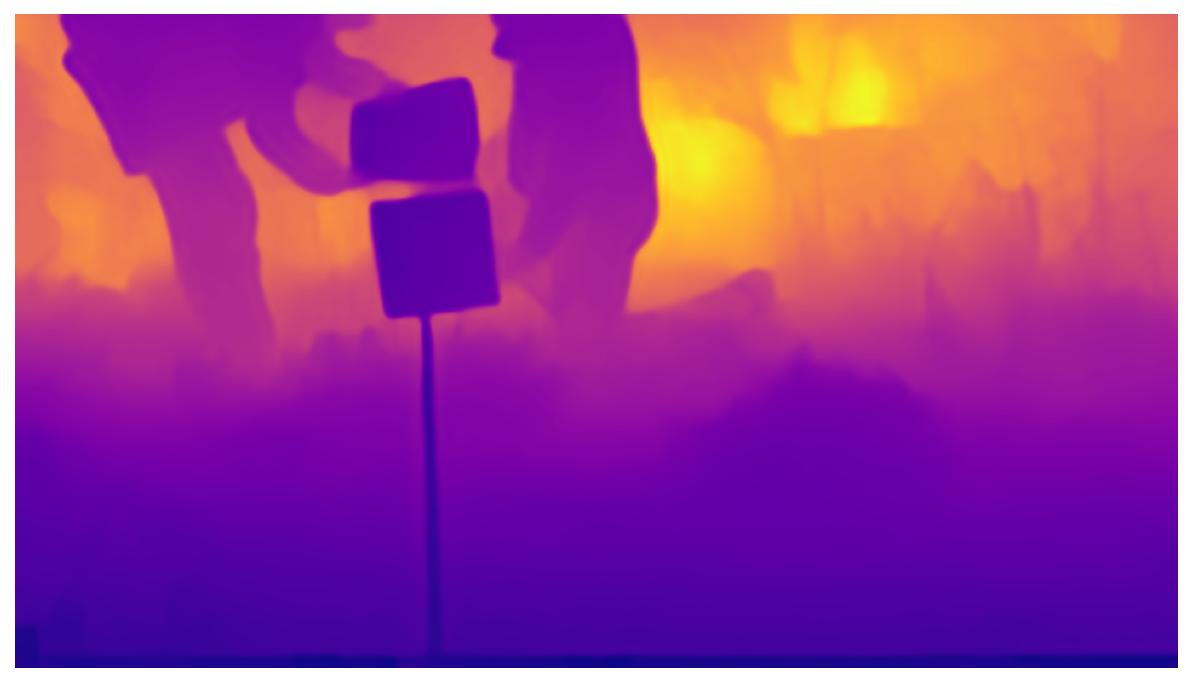}
        \caption{ZoeDepth}
        \label{fig:img4}
    \end{subfigure}
    \hfill
    \begin{subfigure}[b]{0.45\textwidth}
        \centering
        \includegraphics[width=\textwidth]{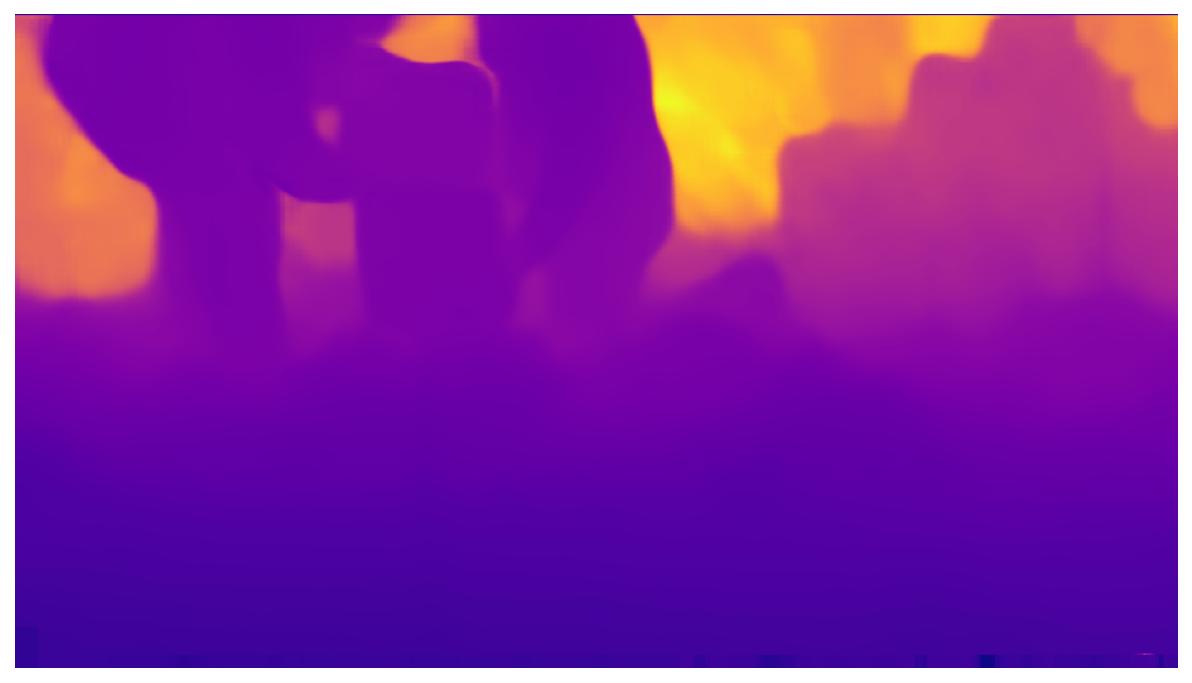}
        \caption{Metric3D}
        \label{fig:img5}
    \end{subfigure}

    \caption{Qualitative comparison of depth estimation methods on a wildlife monitoring scene. (a) Ground truth RGB image showing the outdoor environment with measurement setup. (b-e) Predicted depth maps where warmer colors (yellow/orange) indicate closer distances and cooler colors (purple/blue) represent farther distances.}
    \label{fig:qualitative-results}
\end{figure}

Visual inspection of the predicted depth maps reveals important qualitative differences between methods. Depth Anything V2 (b) produces smooth depth transitions with clear distinction between foreground and background elements, showing good preservation of spatial relationships. ML Depth Pro (c) shows more pronounced contrast between near and far regions, capturing the overall scene structure but with potentially less nuanced depth variations in intermediate distances. ZoeDepth (d) produces depth maps with less spatial coherence and more abrupt transitions, which correlates with its quantitative performance limitations in outdoor natural environments.
Metric3D (e) generates depth maps with similar structural quality and smooth gradients, demonstrating excellent correlation with the expected depth distribution.
All methods successfully identify the general depth structure, but differ in their ability to capture fine-grained depth variations and maintain spatial consistency.

\section{Conclusions}\label{sec:conclusions}

This work presents the first comprehensive benchmark of monocular depth estimation methods for wildlife monitoring applications. Through evaluation on 93 camera trap images, we demonstrate that Depth Anything V2 achieves superior performance with 0.454m mean absolute error, while existing methods like ZoeDepth show significant degradation in natural outdoor environments.

Our key findings include:\begin{itemize}
    \item median-based depth extraction consistently outperforms mean-based approaches,
    \item methods optimized for urban/indoor scenes may fail in wildlife settings
    \item the projection-based baseline provides competitive accuracy (0.505m MAE), validating our evaluation framework.
\end{itemize}

We recommend Depth Anything V2 with median extraction for wildlife applications requiring accurate distance measurements. Future work should focus on developing wildlife-specific datasets, domain adaptation techniques, and multi-modal approaches combining depth estimation with other sensing modalities common in wildlife monitoring.

This benchmark establishes a foundation for method development and provides practical guidance for implementing depth estimation in wildlife monitoring systems, contributing to more effective conservation technologies.

\paragraph{Acknowledgments}
This research is partially supported by the BIG\_PICTURE project, funded as a BIODIVERSA+ Initiative.

\bibliographystyle{plain}
\bibliography{iciap25}

\end{document}